\documentclass[11pt]{article}

\usepackage[final]{acl}
\usepackage{inconsolata}
\usepackage{xspace}
\usepackage{makecell}
\definecolor{shadecolor}{gray}{.9}
\definecolor{Gray}{gray}{0.9}
\newcommand{\modelname}{LogitsCoder\xspace}

\usepackage{times}
\usepackage{latexsym}

\usepackage[T1]{fontenc}

\usepackage[utf8]{inputenc}

\usepackage{microtype}

\usepackage{inconsolata}

\usepackage{graphicx}
\usepackage{algorithm}
\usepackage{algorithmic}
\usepackage{times}
\usepackage{pifont}
\usepackage{latexsym}
\usepackage{booktabs}
\usepackage{graphicx}
\usepackage{subcaption}
\usepackage{enumitem}
\usepackage{amsmath}
\usepackage{multirow}
\usepackage{colortbl}
\usepackage{tabularx}
\usepackage{booktabs}
\usepackage{siunitx}
\usepackage{amssymb}
\usepackage{color}
\usepackage{svg}
\usepackage[T1]{fontenc}
\usepackage[utf8]{inputenc}
\usepackage{microtype}

\usepackage{inconsolata}
\usepackage{graphicx}
\usepackage{multirow}
\usepackage{subcaption}
\usepackage{float}
\usepackage{enumitem}
\usepackage{verbatim}
\usepackage{colortbl}
\usepackage{mdframed}
\usepackage{adjustbox}
\usepackage{tabularx}
\usepackage{enumitem}
\usepackage{tcolorbox}
\usepackage{natbib}
\tcbuselibrary{skins, breakable} 
\newcommand{\task}[1]{Debugging}
\title{\modelname: Towards Efficient Chain-of-Thought Path Search via Logits Preference Decoding for Code Generation }

\author{
Jizheng Chen$^{1}$\thanks{Equal Contribution.}, 
Weiming Zhang$^{1}$\footnotemark[1], 
Xinyi Dai$^{1}$, Weiwen Liu$^{1}$, Kounianhua Du$^{1}$, \\
\bf Yasheng Wang$^{2}$, Ruiming Tang$^{2}$, Yong Yu$^{1}$, Weinan Zhang$^{1}$ \thanks{Corresponding author.}\\
$^{1}$Shanghai Jiao Tong University, $^{2}$Huawei Noah’s Ark Lab \\
Shanghai, China \\
\texttt{\{humihuadechengzhi, WeimingZhang\_2020, wnzhang\}@sjtu.edu.cn}
}

\begin{document}

\maketitle
\begin{abstract}

Code generation remains a challenging task that requires precise and structured reasoning. Existing Test Time Scaling (TTS) methods, including structured tree search, have made progress in exploring reasoning paths but still face two major challenges: (1) underthinking, where reasoning chains tend to be shallow and fail to capture the full complexity of problems; and (2) overthinking, where overly verbose reasoning leads to inefficiency and increased computational costs. To address these issues, we propose \modelname, a novel framework that enhances chain-of-thought reasoning through lightweight, logit-level control mechanisms for code generation. \modelname iteratively generates and refines reasoning steps by first steering token selection toward statistically preferred patterns via Logits Preference Decoding, then selecting and aggregating diverse reasoning paths using Logits Rank Based Path Selection and Thoughts Aggregation. This results in coherent and effective reasoning chains that balance depth and efficiency. Extensive experiments demonstrate that \modelname produces more efficient and higher-quality reasoning chains, leading to superior code generation performance compared to baseline methods.
\end{abstract}


\section{Introduction}

Large language models (LLMs) have demonstrated remarkable success across various domains, while code generation emerging as a particularly challenging task that demands advanced reasoning~\cite{fu2023codeapex, jain2024livecodebench, zhang2025nl}. Early efforts in code generation rely on iterative code-level refinements~\cite{zhang2023self, zhong2024ldb}, which are primarily constrained by limited exploration of reasoning pathways, leading to suboptimal solutions. 

\begin{figure}[h]
    \centering
    
    \includegraphics[width=1.0\linewidth]{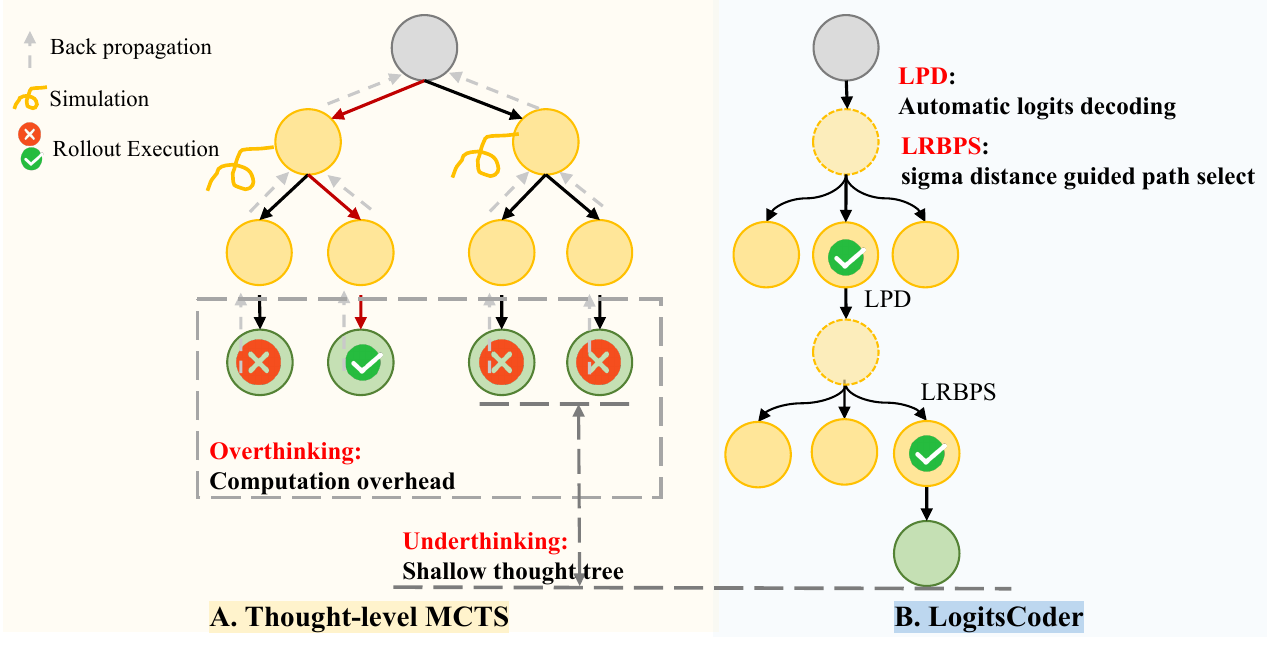}
    
    \caption{(A) Challenges faced by MCTS: shallow search trees causing underthinking, and excessive rollouts leading to overthinking and high computation cost. (B) \modelname framework addressing these challenges with LPD and LRBPS for efficient and deeper reasoning paths.} 
    \label{fig: motivation}

\end{figure}
With the rapid development of reasoning models, Test-Time Scaling (TTS) has emerged as a powerful strategy to stimulate models by expanding thought-level reasoning through approaches like Chain-of-Thought (CoT) prompting~\cite{jaech2024openai, guo2025deepseek, qin2024o1, wang2024openr}. However, current TTS approaches that rely on CoT generation often encounter two failure modes: \textbf{underthinking}, where LLM lacks an accurate understanding of the problem, thus the generated reasoning chains fail to capture problem complexity, and \textbf{overthinking}, where excessively verbose thoughts not only lead to inefficient and convoluted reasoning but can also increase the risk of hallucinations and logical errors, ultimately degrading code generation performance. Both modes highlight the limitations of current CoT-based TTS methods and underscore the need for more robust reasoning strategies to support accurate and efficient code generation.


To address these issues, structured reasoning searching methods have been developed to enhance linear CoT trajectories. One notable approach is Monte-Carlo Tree Search (MCTS)~\cite{zhou2023language}, which incrementally constructs a search tree by systematically expanding, simulating, and evaluating possible reasoning paths to improve decision-making~\cite{zhang2024rest}. Building on this, RethinkMCTS incorporates refinement mechanisms that leverage execution feedback to detect and correct flawed reasoning during the thought searching process~\cite{li2024rethinkmcts}. However, two critical challenges persist, as illustrated in Figure~\ref{fig: motivation}. (1) MCTS often generates \textit{shallow trees} (e.g., depth capped at 4 with 16 rollouts), which restricts comprehensive reasoning for complex code contest problems and leaves underthinking problem unsolved. Deeper and more specific thought steps are required to address complex code generation scenarios. (2) MCTS's iterative rollout process (expand → simulate → backpropagate), incurs \textit{substantial token consumption} and \textit{computational overhead}. This generates redundant reasoning steps, such as repeated thought steps or vague analyses, exacerbating overthinking.

In response, we propose \modelname, a lightweight yet effective framework for efficient CoT path exploration in code generation. As shown in Figure~\ref{fig: motivation}, to overcome shallow reasoning caused by limited MCTS tree depth, \modelname employs Logits Rank Based Path Selection (LRBPS) to sample diverse reasoning trajectories and select the most coherent path using a statistical sigma-distance metric, enabling deeper exploration without expensive tree expansion. To reduce redundant computations from repeated rollouts, we introduce Logits Preference Decoding (LPD), which directly injects token-level preference signals into the decoding process, steering generation toward high-quality reasoning paths. Additionally, the Thoughts Aggregation module combines multiple candidate paths via summarization or selection, guided by lightweight rollout feedback, balancing diversity with factual correctness. Together, these components form a unified framework that improves reasoning depth and efficiency, effectively overcoming both underthinking and overthinking in code generation.

The contributions of our work are summarized as follows: 
\begin{itemize}[leftmargin=10pt]

\item We propose \modelname, a lightweight CoT search framework that enables efficient and effective reasoning. \modelname proposes to apply decoding techniques to Thought Generation and Thought Refinement stages, enhancing reasoning quality while minimizing computational overhead in code generation.

\item We identify underthinking and overthinking as critical limitations in TTS for code generation tasks, and are the first to utilize novel logits decoding methods to effectively address these challenges.

\item We evaluate \modelname on different contest-level tasks, achieving high-quality CoT trajectories with fewer tokens. This demonstrates \modelname's superior efficiency and effectiveness in complex code generation tasks.
 
\end{itemize}

\section{Related Work}
\paragraph{LLMs for Code Generation}

LLMs have demonstrated strong capabilities in code generation, leveraging their ability to model complex linguistic and programming patterns. Existing approaches primarily fall into two categories: one focuses on training-based methods that fine-tune LLMs on extensive code corpora to enhance generation quality~\cite{luo2023wizardcoder, wang2023codet5+, hui2024qwen2}; the other exploits intrinsic LLM reasoning through chain-of-thought (CoT)~\cite{wei2022chain} prompting techniques. While early iterative refinement methods succeeded in improving outputs~\cite{zhang2023self, zhong2024ldb, chen2025debatecoder}, they could not often explore diverse reasoning pathways effectively~\cite{wang2024planning}. More recent plan-based techniques, such as Monte Carlo Tree Search (MCTS) and structured planning, have been proposed to address this limitation by enabling more coherent reasoning exploration~\cite{li2024rethinkmcts, li2024codetree}. Despite these advances, challenges remain in controlling hallucinations and vague generation. To mitigate these, our work introduces a novel logits-guided decoding framework combined with dynamic CoT path search, which jointly improves reasoning accuracy and efficiency in code generation tasks.

\paragraph{LLMs Logits Decoding}
In LLMs, logits are unnormalized output scores reflecting token-level confidence and reasoning tendencies during generation~\cite{zhang2025entropy, wang2025thoughts}. By modifying logits, lightweight control over LLM outputs can be achieved without training, with temperature scaling as the most straightforward approach to adjust output diversity~\cite{wang2025make}. Ongoing efforts focus on regulating logits distributions to guide an effective reasoning process~\cite{tang2024topnsigmalogitsneed}. For instance, Contrastive Decoding enhances output quality by contrasting logits from high- and low-quality generations~\cite{li2022contrastive, ding2024uscd}, while Guided Decoding uses external feedback to steer logits toward task-specific outputs~\cite{agrawal2023monitor}. Recent research shows that logit-induced implicit CoT reasoning outperforms greedy decoding in complex reasoning scenarios~\cite{wang2024chain}. However, dynamically guiding diverse reasoning paths in code generation remains challenging. Our proposed \modelname addresses this gap through dynamic CoT path search, significantly enhancing efficiency and accuracy.

\section{Methodology}
\subsection{Overall Framework}
\begin{figure*}[h]
    \centering
    \includegraphics[width=1.0\textwidth]{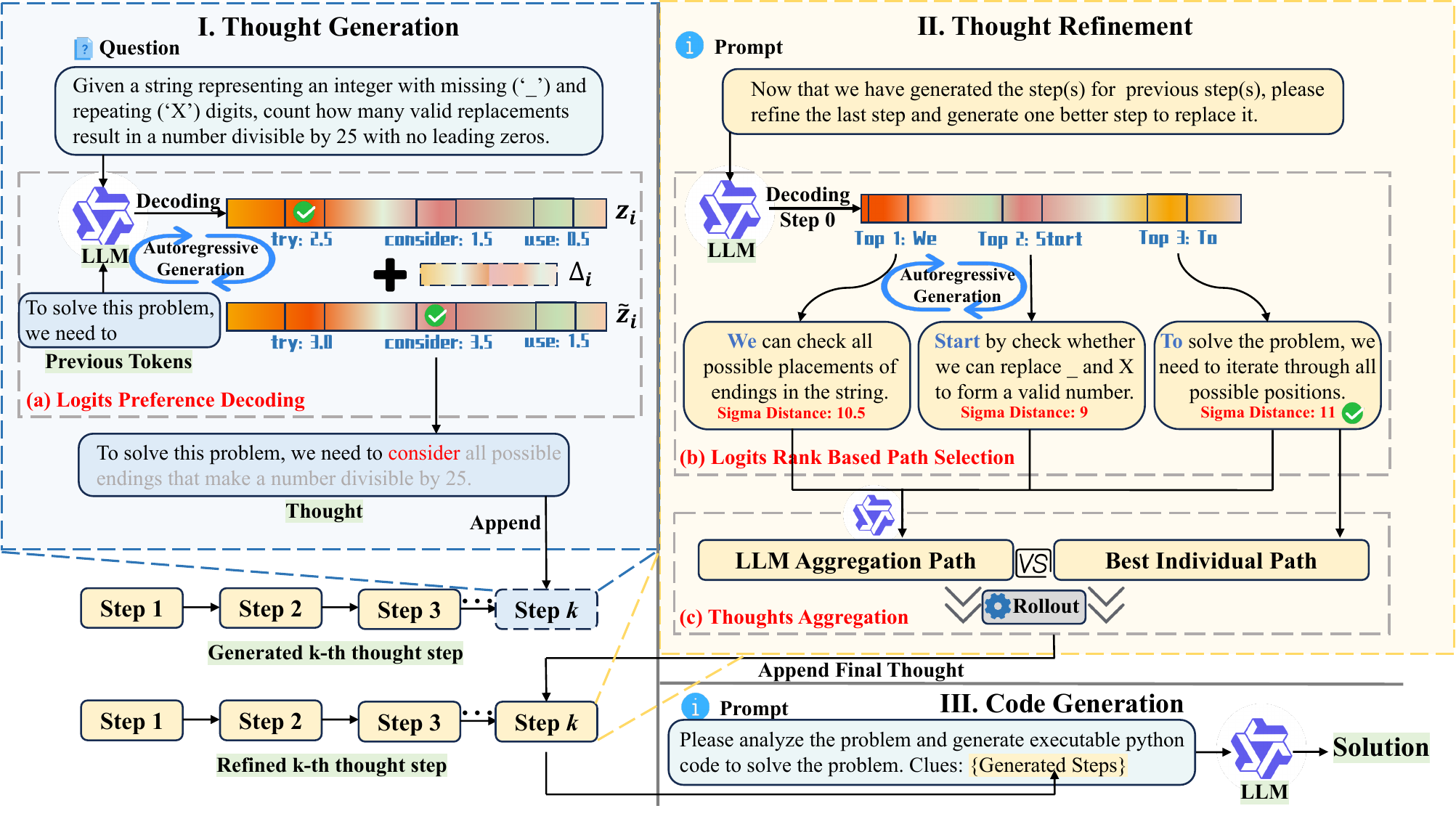}
    
    \caption{Framework overview. \modelname iteratively generates and refines reasoning chains through two stages. In Thought Generation, initial reasoning steps are generated with LPD to bias token selection toward higher-quality outputs. In Thought Refinement, LRBPS and Thoughts Aggregation are applied to enhance step-level accuracy and coherence. This iterative process continues until a complete reasoning chain is formed for Code Generation.} 
    \label{fig:framework}

\end{figure*}

\label{sec:base model}
As shown in Figure~\ref{fig:framework}, our proposed \modelname framework decomposes the code generation task into three key stages: \textbf{Thought Generation stage} uses Logits Preference Decoding (LPD) to produce reasoning steps with higher quality. \textbf{Thought Refinement stage} applies Logits Rank Based Path Selection (LRBPS) and Thoughts Aggregation to iteratively optimize reasoning paths. \textbf{Code Generation stage} generates executable code based on the refined reasoning sequence.

\paragraph{Thought Generation}
Given an input problem description, \modelname generates preliminary reasoning thoughts step by step in natural language using the CoT paradigm. This process employs a novel LPD module to enhance the quality of generated thoughts (More details are provided in Section~\ref{sec: logits preference decoding}). The generation is formulated as a conditional language modeling task:
\begin{equation*}
\label{equ:thought_generation}
p(T|X) = \prod_{i=1}^n p(t_i|X, t_{<i})
\end{equation*}
where $X$ is the input problem description, $T$ is the sequence of thought tokens, $t$ denotes individual tokens during the LLM generation and $n$ is the length of current step. This stage aims to produce reasoning trajectories with more preferred tokens.

\paragraph{Thought Refinement}
In this stage, \modelname refines individual reasoning steps to enhance their correctness, leveraging a proposed LRBPS module and Thoughts Aggregation module (More details are provided in Section~\ref{sec: logits rank based path selection} and Section~\ref{sec: thoughts aggregation}). 
This process is formulated as:
\begin{equation*}
\label{equ:thought_refinement}
p(T'|X,T) = \prod_{i=1}^np_{refine}(t_i|X, t_{<i},T)
\end{equation*}
where $T$ is the current thought step and $T'$ is its refined version, conditioned on the input problem $X$ and prior tokens $t_{<i}$ and current step $T$. 


The refinement module improves each step based on detected logical inconsistencies or chances for improvement. This step-wise process reduces error propagation in the CoT, closely aligns reasoning with the problem, and avoids the high cost of MCTS rollouts.

\paragraph{Code Generation}
Once the refined CoT is fully constructed, the code generation module produces executable code by conditioning on the input problem and the complete sequence of refined thoughts. This process is formulated as:
\[
p(C \mid X, \mathcal{T'}) = \prod_{i=1}^{n} p(c_i \mid X, \mathcal{T'}, c_{<i})
\]
where $\mathcal{T'} = \{ T'_1, T'_2, \dots, T'_m \}$ denotes the refined reasoning chain. 
This enriched context enables \modelname to generate structured, syntactically correct, and semantically accurate code by leveraging the refined reasoning chain.

\subsection{Logits Preference Decoding}
\label{sec: logits preference decoding}

To address underthinking and overthinking in vanilla CoT generation limitations, we propose \textbf{Logits Preference Decoding} (LPD), a mechanism that biases token generation toward sequences indicative of effective problem-solving.


We observe that high-quality CoT sequences display distinct linguistic features, including precise terminology, problem-relevant variables, and logically consistent discourse markers. By statistically comparing positive and negative CoT samples, we extract token-level preference signals that guide the decoding process.

Formally, LPD adjusts output logits during decoding:
\[
\tilde{z}_i = z_i + \Delta_i, 
\quad \Delta_i = f_{\text{pref}}(t_i)
\]
where $z_i$ is the original logit for token $t_i$, and $\Delta_i$ is the preference adjustment computed by $f_{\text{pref}}(t_i)$ based on pre-collected statistical patterns\footnote{$\Delta_i$ is derived by statistically comparing token frequency differences between high-quality and low-quality CoT reasoning steps collected from preliminary CoT experiments. Detailed infomation can be found in Appendix \ref{sec: appendix of LPD}.} from coding tasks. This lightweight adjustment steers the model toward tokens prevalent in high-quality CoT sequences, enhancing reasoning relevance.

LPD is model-agnostic and efficient, enabling effective token-level biasing to explore better CoT paths for refinement and code generation.



\subsection{Logits Rank Based Path Selection}
\label{sec: logits rank based path selection}
While LPD biases the model’s generation towards statistically favorable tokens, it does not fully address the variance across entire CoT trajectories. To mitigate this, we introduce \textbf{Logits Rank Based Path Selection} (LRBPS), which generates diverse reasoning paths upfront and ranks them by token-level statistical consistency.

Concretely, given the input $X$, the model first computes the logits distribution for the initial reasoning token:
\[
z^{(0)} = f_{\text{LM}}(X)
\quad \text{where} \quad
z^{(0)} \in \mathbb{R}^{|V|}.
\]
Instead of greedily sampling the single top token, we sort the logits and select the top-$K$ tokens with the highest probabilities:
\[
\{ t^{(k)} \}_{k=1}^{K} = \text{TopK}(z^{(0)}).
\]
Each of these top-$K$ initial tokens then seeds an independent CoT path. The model independently samples and refines each path step by step:
\[
p(T^{(k)}|X,t^{(k)}) =\prod_{j=2}^{n} p_{refine}(t_i|X,t_{<i}, t^{(k)})
\]
Here $T^{(k)} $ denotes the $k$-th thought path following the fixed $k$-th seed token. This strategy ensures that multiple plausible reasoning directions are explored without requiring heavy rollout-based search as in MCTS.

To select the optimal path, we introduce the \textbf{sigma distance} metric, which evaluates the stability of token logits in each path $T^{(k)}$. High-quality paths exhibit confident and consistent logit distributions, a property rooted in the structure of the logits themselves. Grounded in prior work~\cite{tang2024topnsigmalogitsneed} that analyzes the statistical properties of logits, this metric serves as a direct measure of model confidence; much like how entropy measures uncertainty, a higher sigma distance reflects greater model certainty. For each path, we compute token-level logit scores ${ s_i^{(k)} }$, where $s_i^{(k)}$ is the logit of the $i$-th token. The standard deviation and mean are:
\[
\text{std} = \sqrt{ \frac{1}{n-1} \sum_{i=1}^{n} (s_i - \bar{s})^2 },
\quad
\bar{s} = \frac{1}{n} \sum_{i=1}^{n} s_i.
\]
Next, we compute the probability distribution over these logits and their entropy:
\[
p_i = \text{softmax}(s_i).
\]
Finally, the \textbf{sigma distance} for path $T^{(k)}$ is defined as:
\[
\sigma\text{-distance} = \frac{ \max(s_i) - \bar{s} }{ \text{std} }.
\]

The optimal path is selected as:
\[
T^{*} = \arg\max_{k} \sigma\text{-distance}(T^{(k)}).
\]

By using LRBPS with sigma distance, this method efficiently selects coherent CoT paths, and when combined with Thoughts Aggregation, it enhances the robustness of code generation.


\subsection{Thoughts Aggregation}
\label{sec: thoughts aggregation}
While LPD and LRBPS generate high-quality reasoning paths, individual paths may capture only partial problem perspectives. To enhance robustness, we introduce Thoughts Aggregation, which combines multiple CoT paths into a unified reasoning chain.

Our aggregation employs two approaches:
\paragraph{Path Summarization} 
All candidate paths $T^{(k)}$ and their sigma distance scores $S^{(k)}$ are fed into an aggregation module. An LLM summarizes and refines these paths:
\[
T_{\text{agg}} = \text{Aggregator}\Big( \{ (T^{(k)}, S^{(k)}) \}_{k=1}^{K} \Big).
\]
The model identifies overlapping insights, resolves inconsistencies, and distills a coherent reasoning chain, leveraging diverse perspectives from multiple paths.
\paragraph{Best-Path Selection}
Alternatively, the path with the highest sigma distance is selected directly: 
\[
T_{\text{best}} = \arg\max_{k} \sigma\text{-distance}(T^{(k)})..
\]
This ensures computational efficiency when summarization is redundant, retaining a highly coherent reasoning chain.

To select the final path, both $T_{\text{agg}}$ and $T_{\text{best}}$ undergo a lightweight validation process via rolling out (e.g., partial solution execution):
\[
\text{Score}_{\text{val}}(T_{\text{agg}}), \quad 
\text{Score}_{\text{val}}(T_{\text{best}}).
\]
The path with the higher score, based on execution correctness, is chosen:
\[
T_{\text{final}} = 
\begin{cases}
T_{\text{agg}} & \text{if } \text{Score}_{\text{val}}(T_{\text{agg}}) > \text{Score}_{\text{val}}(T_{\text{best}}) \\[6pt]
T_{\text{best}}& \text{otherwise}.
\end{cases}
\]

By integrating diverse reasoning and practical validation, Thoughts Aggregation produces a robust reasoning chain, enhancing the quality of subsequent code generation.

\begin{table*}[h!]

\renewcommand{\arraystretch}{1.4}

\centering
\resizebox{\textwidth}{!}{
\begin{tabular}{llcccccccccc}
\toprule
 && \multicolumn{6}{c}{\textbf{APPS}} &  \multicolumn{4}{c}{\textbf{CodeContest }} \\
\multicolumn{2}{c}{\textbf{Model}}& 
\multicolumn{3}{c}{Pass Rate} & \multicolumn{3}{c}{Pass@1} & \multicolumn{2}{c}{Pass Rate} & 
\multicolumn{2}{c}{Pass@1}\\
\cmidrule(lr){3-5} \cmidrule(lr){6-8} \cmidrule(lr){9-10} \cmidrule(lr){11-12}
 & & Intro. & Inter. & Comp. & Intro. & Inter. & Comp. & Basic & Advanced & Basic & Advanced \\
\midrule
\textbf{Base}& Zero-shot & 0.3802 & 0.3157 & 0.1950 & 0.1800 & 0.1200 & 0.0700 & 0.3438 & 0.2649 & 0.1918 & 0.1395 \\
\hline
\multirow{3}{*}{\textbf{Reflection-based
Models}} 
 & self-play & 0.5146 & 0.4876 & \underline{0.3017} & 0.3100 & 0.2100 & \underline{0.1700} & 0.4356 & 0.4459 & 0.2877 & \underline{0.3023} \\
 & Reflexion & 0.4538 & 0.4133 & 0.2450 & 0.2700 & 0.2000 & 0.1200 &0.3432  &0.2793  &0.2055  &0.1628  \\
 & RAP & 0.4182 & 0.3632 & 0.2483 & 0.2400 & 0.1300 & 0.0700 &0.2954  &0.2056  &0.1644  &0.1163 \\
 & LDB & 0.4337 & 0.3761 & 0.2367 & 0.2400 & 0.2700 & 0.1000 &0.4139  &0.3087  &0.2740  &0.1628  \\
 \hline
\multirow{3}{*}{\textbf{Search-guided  Reasoning Models}} 
 & LATS & 0.5321 & 0.4442 & 0.2386 & \underline{0.3500} & 0.2100 & 0.1300 &0.4412  &0.3351  & 0.2603 &0.1860  \\
 & MCTS & 0.5627 & 0.5667 & 0.2733 & 0.3200 & 0.2300 & 0.1300 &0.4114  & \underline{0.4590} &0.2329  &0.2326  \\
 & RethinkMCTS & \underline{0.5995} & \underline{0.5888} & 0.2700 & 0.3100 & \underline{0.2900} & 0.1300 &0.4003  & 0.4238 &0.2381  &0.2587  \\
 \hline
\multirow{2}{*}{\textbf{Decoding-based Models}} 
 & Contrastive Decoding & 0.5623 & 0.5463 & 0.2883 & \underline{0.3500} & 0.2500 & 0.1300 &\underline{0.4512}  & 0.4286 &\underline{0.2883}  &0.2466  \\
  &Guided Decoding & 0.5615 & 0.5527 & 0.3150 & 0.3200 & 0.2500 & \underline{0.1700} &0.4360  & 0.3996 &0.2603  &0.2326  \\
 \rowcolor{Gray}
 & \modelname(Ours) & \textbf{0.6221} & \textbf{0.6130} & \textbf{0.3500} & \textbf{0.3800} & \textbf{0.3400} & \textbf{0.2000} & \textbf{0.4999} & \textbf{0.5040} & \textbf{0.3151} & \textbf{0.3256} \\

\bottomrule
\end{tabular}}
\caption{Major results. For each backbone, the best result(s) are marked in bold, and the second best result(s) are underlined. Each method is run over three times to get the best result for a fair comparison.}
\label{tab: main table}
\end{table*}

\begin{table*}[]

\resizebox{\textwidth}{!}{%
\begin{tabular}{cccccccccccccccc}
\toprule
\multirow{3}{*}{\textbf{Method}} & \multicolumn{9}{c}{\textbf{APPS}}                                                                           & \multicolumn{6}{c}{\textbf{CodeContest}}                                                \\
& \multicolumn{3}{c}{Input (k tokens)} & \multicolumn{3}{c}{Output (k tokens)} & \multicolumn{3}{c}{Efficiency (k tokens)} & \multicolumn{2}{c}{Input (k tokens)} & \multicolumn{2}{c}{Output (k tokens)} & \multicolumn{2}{c}{Efficiency (k tokens)} \\
\cmidrule(lr){2-4} \cmidrule(lr){5-7} \cmidrule(lr){8-10} \cmidrule(lr){11-12} \cmidrule(lr){13-14} \cmidrule(lr){15-16} 

                                 & Intro.     & Inter.     & Comp.     & Intro.      & Inter.     & Comp.     & Intro.       & Inter.       & Comp.      & Basic             & Adv.            & Basic             & Adv.             & Basic               & Adv.               \\
\midrule
RethinkMCTS                      & 13072      & 12900      & 15163     & 1922        & 1966       & 2414      & 282          & 286          & 740        & 12311             & 8184            & 1766              & 1166             & 396                 & 248                \\
Base+Decoding+Best               & 10187      & 9130       & 10672     & 971         & 129        & 1369      & 201          & 196          & 430        & 11958             & 6068            & 1291              & 649              & 331                 & 174                \\
Base+Decoding+Agg.               & 12748      & 11355      & 12287     & 924         & 1037       & 1420      & 253          & 237          & 422        & 9700              & 6113            & 948               & 630              & 261                 & 173                \\
\modelname                      & 12835      & 12777      & 12403     & 1140        & 207        & 1733      & 243          & 248          & 453        & 11561             & 6068            & 124               & 649              & 250                 & 158        \\       \bottomrule
\end{tabular}%
}
\caption{Token usage and efficiency ratio for different methods on APPS and CodeContest datasets.}
\label{tab:token}
\end{table*}

\section{Experiment}
In this section, a series of experiments are conducted to answer the following research questions(RQs):
\begin{itemize}[leftmargin=27pt]
    \item[\textbf{RQ1}] How does our proposed \modelname perform against the baselines?
    \item[\textbf{RQ2}] Does the proposed \modelname perform an efficient and lightweight CoT path search?
    \item[\textbf{RQ3}] Does each component of \modelname bring a performance gain?
    \item[\textbf{RQ4}] How is the performance of \modelname in terms of test time scaling?
    \item[\textbf{RQ5}] What is the specific impact of our method on the model's reasoning chain?
\end{itemize}
\subsection{Experimental Setup}
\subsubsection{Datasets and Metrics}
Our evaluation encompasses two widely adopted datasets: APPS~\cite{hendrycks2021measuring} and CodeContest~\cite{doi:10.1126/science.abq1158}.
\begin{itemize}[leftmargin=10pt]
    \item \textbf{APPS} contains three levels of difficulty: introductory, interview, and competition. We evaluate all the methods on the formal 100 problems of each difficulty.
    \item \textbf{CodeContest} categorizes the test problems into basic and advanced levels on their difficulty tags, containing 73 and 43 problems, respectively.
\end{itemize}

we use \textit{pass rate} and \textit{pass@1} as evaluation metrics for code generation accuracy. The \textit{pass rate} reflects the average percentage of private test cases that the generated programs pass across all problems. In contrast, \textit{pass@1} indicates the proportion of problems for which the generated programs successfully pass all private test cases.


\subsubsection{Baselines}
We compare our proposed \modelname with a series of competitive methods to validate the effectiveness of \modelname, including \textbf{self-reflecion-based methods}: self-play\citep{madaan2022language}, Reflexion\cite{shinn2024reflexion}, RAP\citep{kagaya2024rap}, \textbf{code debugging methods} LDB\cite{zhong2024debug}, \textbf{search-guided reasoning methods} MCTS\citep{zhu2022core}, RethinkMCTS\citep{li2024rethinkmcts}, LATS\citep{zhou2023language}, along with two \textbf{decoding-based methods}: Contrastive Decoding\citep{li2023contrastivedecodingopenendedtext} and Guided Decoding\citep{wang2024chainofthoughtreasoningprompting}.

\subsubsection{Implementation}
We select Qwen2.5-14B-Instruct~\cite{yang2024qwen2} as the backbone, which is a powerful and representative LLM. Closed-source models like GPT-4~\cite{jaech2024openai} are excluded because our method requires real-time token-level logits manipulation, which current APIs do not support. For fair comparison, all baselines and \modelname use up to 20 code execution rollouts. Following prior work~\citep{li2024rethinkmcts,chen2022codet}, we split test cases evenly into public sets for execution feedback and private sets for final evaluation.

\subsection{Overall Performance (RQ1)}

In this section, we compare our proposed \modelname with baseline methods for code generation, with results presented in Table~\ref{tab: main table}. From these results, we can observe that (1) \modelname achieves the best performance across all settings on both datasets, demonstrating superior CoT quality and code correctness. (2) Search-guided methods, such as MCTS and RethinkMCTS, outperform reflection-based approaches. This is because they enable broader and more structured exploration of reasoning paths. However, \modelname surpasses these baselines with a lightweight strategy that efficiently generates and selects high-quality reasoning paths, demonstrating the efficacy of targeted, computationally efficient reasoning. (3) While decoding-based methods improve CoT quality by adjusting token-level probabilities, \modelname consistently outperforms them by extending control from individual tokens to the entire reasoning trajectory, enabling more coherent and globally optimized chain-of-thought paths.



\begin{figure*}[h]
    \centering
    
    \includegraphics[width=0.95\textwidth]{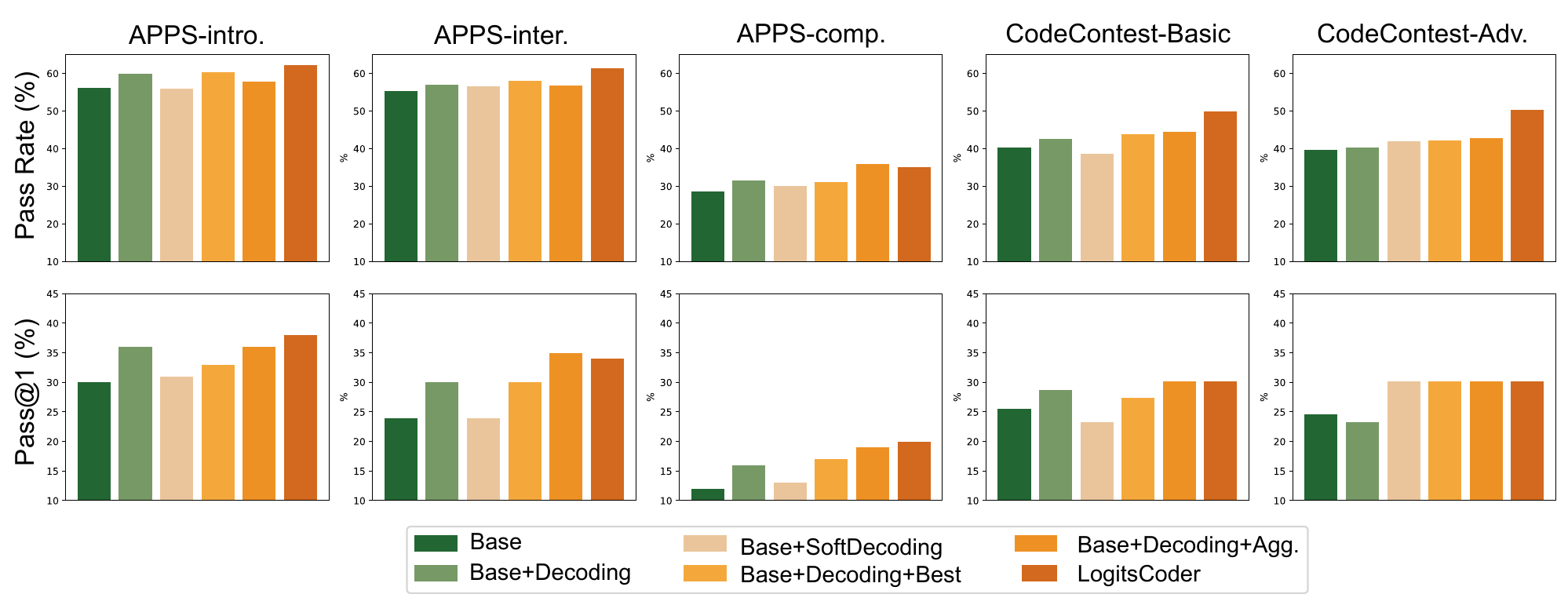}
    
    \caption{Ablation study: Performance of variants of \modelname on APPS and CodeContest datasets.} 
    \label{fig:ablation}
    \vspace{-10pt}
\end{figure*}
\begin{figure}[h]
    \centering
    
    \includegraphics[width=1.0\linewidth]{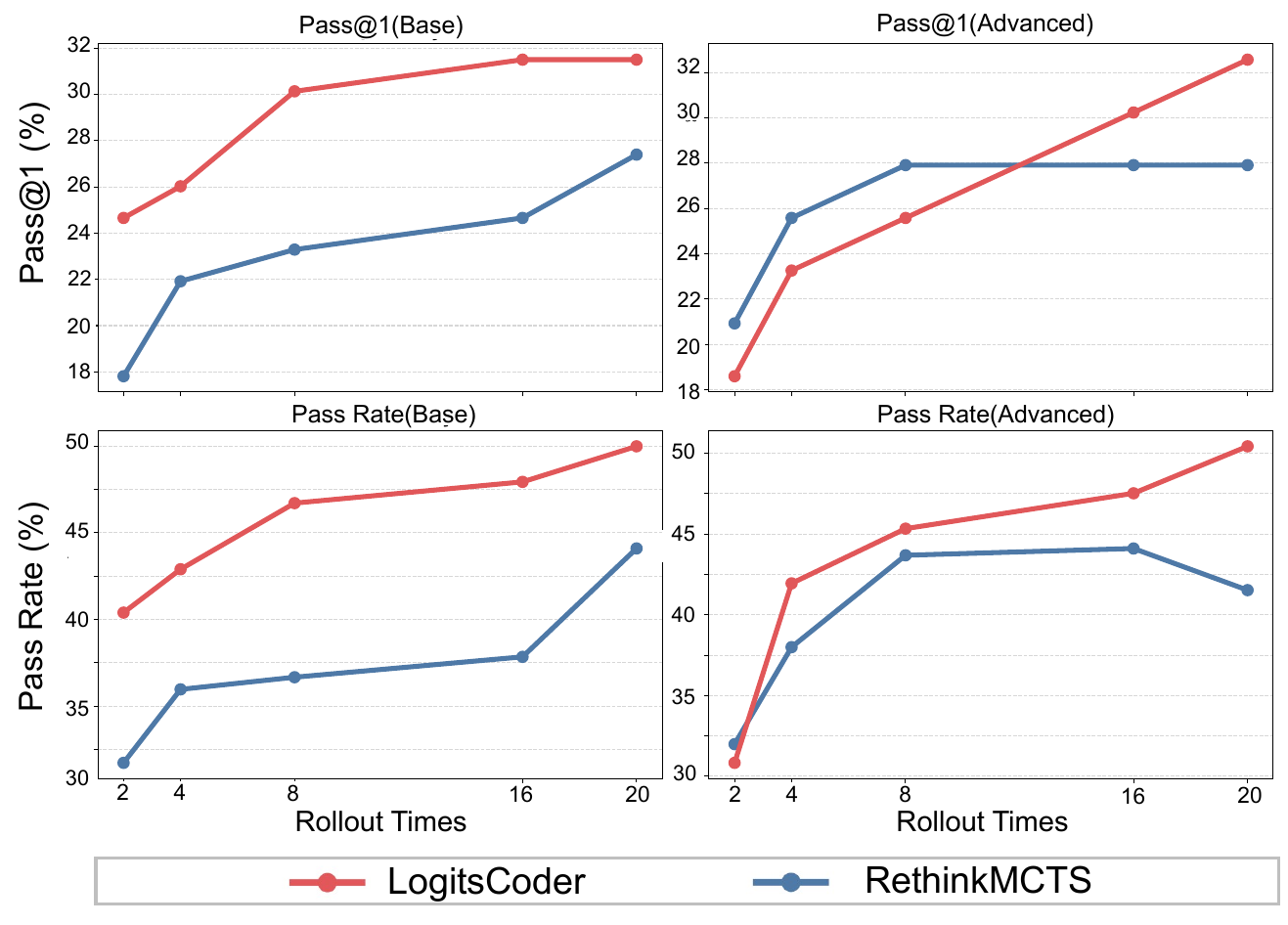}
    
    \caption{TTS performance between \modelname and RethinkMCTS, with rollout times set from 2 to 20. } 
    \label{fig:bar}
    \vspace{-20pt}
\end{figure}
\subsection{Token Efficiency (RQ2)}

To evaluate thought space search efficiency, we compare \modelname with two simplified variants:
\begin{itemize}[leftmargin=10pt]
    \item \textbf{Base+Decoding+Agg.} performs thought generation and refinement with LPD and aggregates reasoning paths via large-model summarization.
    \item \textbf{Base+Decoding+Best} similarly generates and refines thoughts but selects the single best path based on sigma-distance without summarization.
\end{itemize}
Both exclude the final dynamic rollout-based selection in full \modelname to isolate component effects. We compare these methods against the search-based baseline RethinkMCTS under the same rollout budgets. We use input and output token consumption as core indicators and introduce an \textbf{Efficiency} metric as: 


\[
   \text{Efficiency} = \frac{\text{Input Tokens} + 2 \times \text{Output Tokens}}{\text{Pass Rate}}
\]
The 2x weight for output tokens reflects their higher computational cost, a convention consistent with both established practices in large-scale transformer inference~\cite{pope2022efficientlyscalingtransformerinference, shazeer2019fasttransformerdecodingwritehead} and the pricing models of commercial LLM APIs.


\begin{figure*}[h]
    \centering
    
    \includegraphics[width=0.95\textwidth]{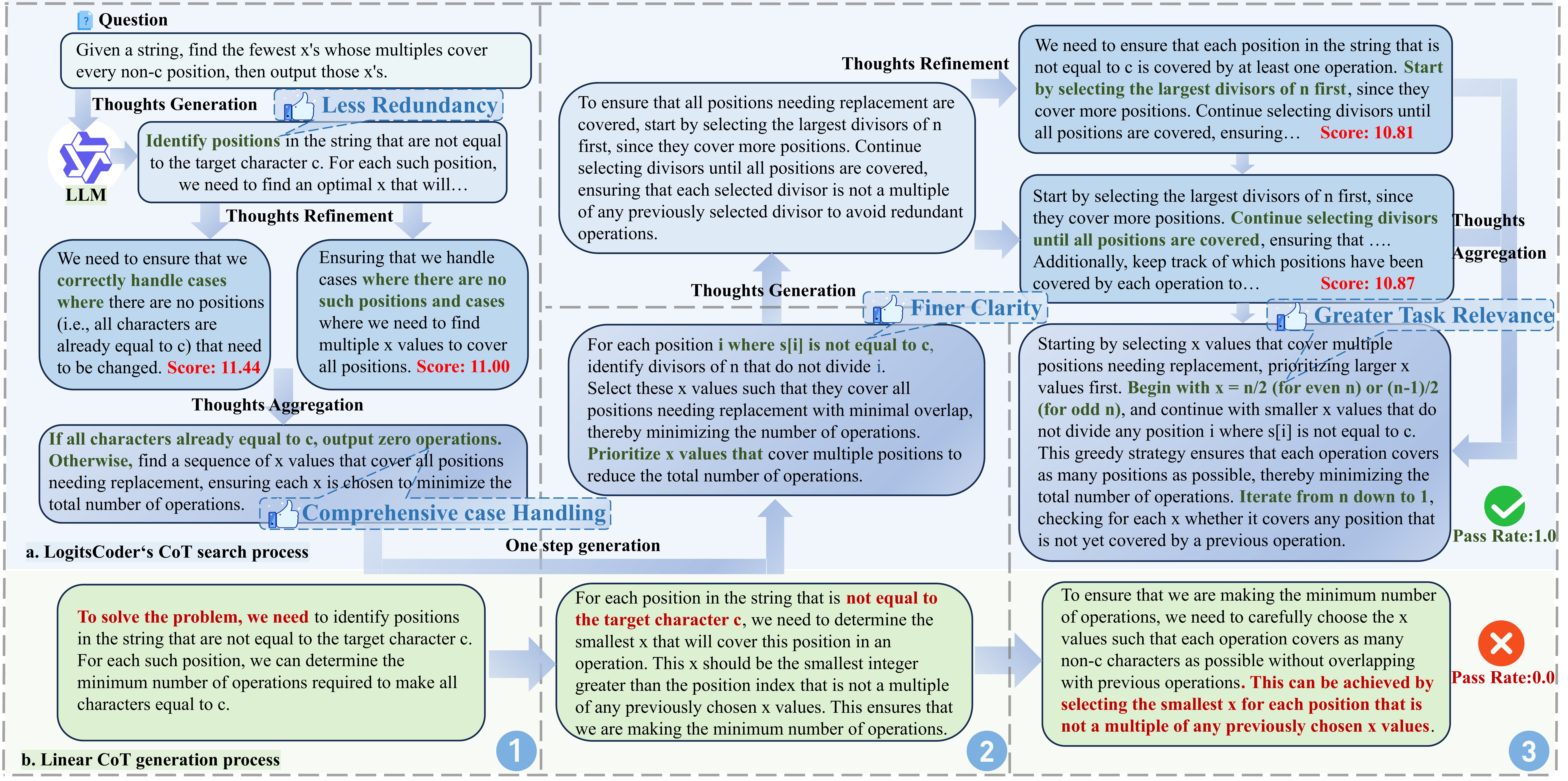}
    
    \caption{Case study of \modelname's search process and Linear CoT generation process on a programming task.} 
    \label{fig:case study}
    \vspace{-10pt}
\end{figure*}

As shown in Table~\ref{tab:token}, all \modelname variants use significantly fewer tokens than RethinkMCTS to achieve comparable or better Pass Rates on both datasets. Notably, \modelname attains the highest accuracy with lower token usage, demonstrating that our lightweight logits-guided search explores the reasoning space more efficiently while maintaining state-of-the-art performance.
\subsection{Ablation Study (RQ3)}



To evaluate the contribution of each component in \modelname, we perform ablation studies with incremental module additions. Figure~\ref{fig:ablation} shows Pass@1 and Pass Rate for each variant on APPS and CodeContest datasets.

Starting from the \textit{Base}, which includes sequential thought generation, refinement, and code generation without path search or decoding adjustments. We add LPD in \textit{Base+Decoding} to improve reasoning via token-level preference signals. \textit{Base+SoftDecoding} uses a simplified version of LPD, where preference weights are derived from statistical averages rather than being explicitly learned. To control reasoning paths, \textit{Base+Decoding+Best} applies LRBPS to pick the most coherent path based on sigma-distance. In contrast, \textit{Base+Decoding+Agg} replaces this with Thoughts Aggregation to combine multiple candidate paths through large-model summarization. The full \modelname integrates all modules into a unified framework.

From the results, we observe the following:
\begin{itemize}[leftmargin=10pt]
    \item \textbf{Decoding modules} (both learned and statistical) provide clear improvements over the base model, showing that token-level preference control is essential for better reasoning path search.
    \item Both \textbf{best-path selection} and \textbf{path summarization} independently enhance performance. Best-path selection boosts Pass@1 and Pass Rate by choosing the most coherent path, while LLM summarization further improves results by integrating multiple paths.
    \item Combining the two with dynamic \textbf{thoughts aggregation}, which selects between aggregated and best paths via rollout feedback, achieves the highest overall performance, especially on complex problems.
    
\end{itemize}

These results confirm that each module in \modelname makes a meaningful contribution, and their combination leads to substantial performance gains over the base model and all ablation variants.

\subsection{Test Time Scaling Performance (RQ4)}


We conduct experiments on TTS performance to address two key questions: (1) As we increase the number of rollouts, can \modelname continuously improve its performance? (2) Does \modelname possess TTS capability comparable to or better than RethinkMCTS?

To answer these questions, we vary the number of rollouts (2, 4, 8, 16, 20) for both \modelname and RethinkMCTS, and evaluate them on the CodeContest dataset for both basic and advanced problems. The following conclusions can be drawn:
(1) \modelname demonstrates robust TTS on CodeContest as rollouts increase from 2 to 20, reaching Pass Rate scores of 49.99\% (easy) and 50.44\% (hard). This reflects its efficient exploration of high-quality CoT paths and enhanced code generation.
(2) \modelname shows a stronger improvement trend as rollout counts increase. Although the performance gap fluctuates at lower rollout numbers, \modelname ultimately surpasses RethinkMCTS with higher Pass Rate and Pass@1 at more rollouts. This demonstrates \modelname’s better ability to leverage computational budget for superior test-time scaling performance efficiently.


\subsection{Case Study (RQ5)}

In this section, we conduct a qualitative case study by comparing the reasoning chains produced by \modelname with those generated by conventional linear CoT models on a representative CodeContest problem. As shown in Figure~\ref{fig:case study}, \modelname produces a CoT with better quality. Specifically:

\begin{itemize}[leftmargin=10pt]
    \item \textbf{Less Redundancy:} \modelname avoids verbose or repetitive statements that are commonly present in vanilla CoT models. Generic remarks or redundant summaries are replaced by focused problem decomposition steps.
    \item \textbf{Comprehensive Case Handling:} Our method systematically considers special and edge cases, such as when all characters are already the target $c$, or when certain positions require no change.
    \item \textbf{Finer Granularity and Clarity:} The reasoning steps produced by \modelname are more detailed and better organized, breaking down the problem into clear sub-tasks. 
    \item \textbf{Greater Task Relevance:} Every step in \modelname’s output directly contributes to the overall solution, whereas the baseline model may generate steps that are either irrelevant or too abstract to be actionable.
\end{itemize}









\section{Conclusion}
In this paper, we propose \modelname, a novel framework that replaces the traditional MCTS for thought search with lightweight logit-level modules. Our method efficiently generates and iteratively refines chain-of-thought reasoning paths, achieving higher-quality reasoning with significantly fewer tokens. Extensive experiments demonstrate \modelname's advantages in both reasoning efficiency and code generation accuracy.

\section*{Limitations}
Despite the substantial performance improvements demonstrated by \modelname, this work has several limitations that warrant future investigation:
\paragraph{Limited Generalizability of Experimental Evaluation} Our experiments focus on the competetive benchmarks, which consist primarily of self-contained, algorithmic problems typical of programming competitions, following previous works\citep{li2024rethinkmcts,chen2025debatecoder}. While these datasets effectively showcase \modelname’s strengths in complex algorithmic reasoning, they do not fully reflect the challenges of real-world software development—such as managing complex dependencies and interacting with external libraries. More realistic benchmarks present these challenges, remain unexplored in our study. 

\paragraph{Under-Exploration on More Open-Source Base Models} We chose Qwen2.5-14B-Instruct as our underlying reasoning model due to its open-source nature and support for real-time logits access, whereas closed-source state-of-the-art models (e.g., GPT-4, Gemini, Claude) do not currently permit this. However, the overall effectiveness of any reasoning framework is heavily influenced by the base model’s intrinsic capabilities. It remains an open question whether \modelname’s explicit guidance—through statistical priors and logits-stability mechanisms—would still yield significant marginal improvements when applied to more capable models.

These limitations highlight areas where LogitsCoder can evolve. We believe that addressing them could further enhance the framework’s adaptability and effectiveness across various structured reasoning tasks.
\bibliography{custom}

\appendix

\section*{Appendix}
\section{Dataset Details}
We assess the performance of our \modelname framework and all the baselines using two commonly used code generation datasets: APPS~\citep{hendrycks2021measuring} and CodeContest~\citep{doi:10.1126/science.abq1158}. The APPS dataset contains 5000 programming challenges split into three difficulty categories—introductory, interview, and competition—with 5000 problems for training and another 5000 for testing. The CodeContest dataset, sourced from the online competitive programming platform, includes problems of various difficulty levels, categorized by "ratings". For evaluation, we focus on problems rated below and above 15, dividing them into basic and advanced sets.

For APPS dataset, we select 100 problems from each difficulty level to ensure a balanced evaluation, and for CodeContest dataset, we have 73 problems for basic and 43 problems for advanced set. Each dataset uses the same set of public test cases for algorithm optimization and private test cases for performance evaluation. 

\section{Baseline Details}

To give a thorough insight into our experimental setup, we provide an in-depth description of each baseline code generation method used for comparison with \modelname. All baselines utilize code execution feedback to progressively refine the code.

\paragraph{Self-play~\citep{madaan2022language}}This method improves the output of LLMs by using an iterative feedback and refinement process. After an initial output is generated by the model, the same model provides feedback on its own output. This feedback is then used to refine the output, repeating the process for multiple iterations until a stopping condition is met. This method does not require additional training or external supervision, instead relying on a single LLM to generate, critique, and refine its own outputs. 
\paragraph{Reflexion~\citep{shinn2024reflexion}} Reflexion enables language agents to improve through code execution traces. After each trial, the agent generates natural language feedback about its own mistakes and stores it in memory. This feedback is then used to guide future decisions, acting as a semantic signal for learning. 
\paragraph{RAP~\citep{kagaya2024rap}}RAP enhances language agents by combining retrieval-augmented memory with task planning. The agent retrieves relevant past trajectories through code execution from an external memory based on the current programming problem, then conditions its planning and actions on this retrieved experience.
\paragraph{LDB~\citep{zhong2024debug}} LDB breaks down programs into fundamental code blocks and conducts execution-level analysis on each one. Its reasoning approach is detailed and structurally informed, yet remains confined to the scope of individual code blocks.
\paragraph{LATS~\citep{zhou2023language}} LATS unifies reasoning, acting, and planning in language models by integrating Monte Carlo Tree Search (MCTS) with large language models (LLMs). In LATS, the LLM acts as both an agent and a value function. LATS samples multiple candidate solutions, evaluates them using test feedback (e.g., assert statements), and selects the best-performing one via tree search. 
\paragraph{MCTS~\citep{zhu2022core}}Monte Carlo Tree Search (MCTS) treats code generation as a thought-level sequential decision-making problem, where each thought step corresponds to an action. It incrementally builds a tree of partial CoT sequences, simulating code completions to evaluate possible next thought steps. Using techniques like UCT to balance exploration and exploitation, MCTS selects promising paths and updates node values based on code quality metrics or execution results. 
\paragraph{RethinkMCTS~\citep{li2024rethinkmcts}}RethinkMCTS introduces a "rethink" operation, where erroneous thoughts are corrected based on this feedback, improving search quality and efficiency, to the MCTS process. 

RethinkMCTS enhances CoT reasoning via Monte Carlo Tree Search with path-level rethink operations, but incurs high token and computation costs; in contrast, \modelname achieves efficient CoT exploration through logits-based decoding, ranking, and aggregation without full tree traversal.
\paragraph{Contrastive Decoding~\citep{li2023contrastivedecodingopenendedtext}}
 Contrastive Decoding (CD) generates text by selecting tokens that maximize the difference between the log-likelihoods derived from two logits, denoted as $z^{+}$ and $z^{-}$, while simultaneously enforcing a plausibility constraint to maintain naturalness and fluency. Specifically, we define $z^{+}$ as the statistical distribution of tokens obtained from problems with satisfactory performance (accuracy $> 0.5$), and $z^{-}$ as the corresponding distribution from problems with unsatisfactory performance (accuracy $\leq 0.5$).
\paragraph{Guided Decoding~\citep{wang2024chainofthoughtreasoningprompting}}Guided decoding steers code generation by incorporating guidance function into the token selection process during inference to adjust the model's logits distributions. In this paper, for a fair comparison, we use sigma distance as the guidance function.


\begin{table}[h]
\centering
\begin{tabular}{p{8cm}}
\toprule
\textbf{High-Quality Words (Increase Logits)} \\
\midrule
cannot, verify, rest, desired, yes, met, remember, actual, per, optimizing, best, simulate, its, observe, even, resulting, point, applies, lengths, matches, conditional, allowed, respective, optimized, p\_1, top, adjacent, print, time, character, therefore, firstly, configuration, vertical, corresponding, redundant, functions, works, choosing, otherwise, running, insight, consist, length, arrangement, maintainability, substrings, upper, effect, product, program, allow, exactly, storing, maximizing, nested, quick, parity, relationship, enough, keeping, identical, compare, three, into, t, zeros, processing, readability, encountered, horizontal, covered, verifying, sure, always, checks, distance, table, part, match, choose \\
\midrule
\textbf{Low-Quality Words (Decrease Logits)} \\
\midrule
placing, meets, unique, begin, especially, difference, programming, every, dynamic, identify, ways, calculation, element, filling, largest, defining, requirements, further, structures, specifically, compute, processes, strategy, been, down, less, help, those, valid, maintaining, correctly, outputting, operations, move, optimal, construct, positive \\
\bottomrule
\end{tabular}
\caption{Logits Adjustment for High-Quality and Low-Quality Words.}
\label{table: lpd}
\end{table}

\section{Details of Logits Preference Decoding}
\label{sec: appendix of LPD}
LPD is designed to address the issues of underthinking and overthinking in code generation, which we identify in chain-of-thought (CoT) reasoning. It works by guiding the decoding process of large language models (LLMs) using token-level statistical preference signals, ensuring that the generated reasoning follows effective problem-solving patterns.

To clarify the data source and address reproducibility concerns, the statistical priors used by the Logits Preference Decoding (LPD) module are derived from the Base model’s preliminary CoT experiments on the CodeContest dataset, specifically combining results from the easy and hard subsets used in Section 4.4 (Ablation Study). For each problem, we collected CoT reasoning traces and labeled their outcomes according to execution accuracy. High-quality samples are defined as those achieving accuracy greater than 0.5, while low-quality samples are those below this threshold.

This merged dataset ensures both generality and scalability: generality, because the preference statistics are extracted from diverse tasks across different difficulty levels and can be reused across experiments without re-collection; and scalability, because these statistics have been empirically verified to transfer effectively to other datasets and reasoning domains. Once computed, the same statistical preferences can be directly applied to new inference tasks, providing a reusable calibration signal.

During token generation, LPD adjusts the logits for each token based on these pre-collected statistical patterns observed in high-quality code generation sequences. This adjustment steers the decoding process toward higher-quality reasoning paths, improving the coherence and accuracy of generated code. Table~\ref{table: lpd} lists representative examples of high-quality tokens whose logits are increased and low-quality tokens whose logits are decreased.

\section{Prompts}
\label{pt:prompts} %

We present prompts that represent the entire framework’s operation process, including steps for thought generation, thought refinement, thoughts aggregation and code generation.

\subsection{Prompt for Thought Generation}
\begin{tcolorbox}[colback=white!95!gray, colframe=black, width=1\textwidth, arc=4mm, boxrule=0.5mm]
You will be given a problem desciption.
Please analyze the problem and generate step-by-step clues of how to solve the problem with executable python code. The problem:\\
$\{\text{Problem Description}\}$\\
An example of step-by-step clues for finding the longest palindrome substring is displayed below:\\
$\{\text{Example thought step}\}$\\
You can refer to the above example and generate step-by-step clues for the given problem.\\
Now, please firstly generate the first clue for step 1. (If no step has been generated.)\\
Now that we have generated the clue(s) for previous step(s). Please follow the clues and generate one clue for the next Step {}. (If previous step have been generated.)
\end{tcolorbox}

\vspace{+5pt}

\subsection{Prompt for Thought Refinement}
\begin{tcolorbox}[colback=white!95!gray, colframe=black, width=1\textwidth, arc=4mm, boxrule=0.5mm]
Now that we have generated the clue(s) for previous step(s), please refine the last clue of Step {} and generate one better clue to replace it. Current steps:\\
$\{\text{Current Thought Steps}\}$\\
Here is the **current code**.\\
$\{\text{Current Code Implementation}\}$\\
Here is the **current execution feedback**:
$\{\text{Code Execution Feedback}\}$\\
Please wrap your response into a JSON object that contains keys `Clue of Step {}`.\\
An example is given below. Please format your json output in a **LIST** form.
$\{\text{Example thought step}\}$
\end{tcolorbox}

\vspace{+5pt}

\subsection{Prompt for LLM Thoughts Aggregation}
\begin{tcolorbox}[colback=white!95!gray, colframe=black, width=1\textwidth, arc=4mm, boxrule=0.5mm]
Now that we have generated the clue(s) for previous step(s), we have some candidate clues along with their confidence core to improve clue of Step {} to make it more concise and detailed. Please give me **one** better clue that summarize the candidate clues and contains detailed instruction on solving the problem.\\
Here is the **candidate clues**:
$\{\text{Candidate Step with Sigma Distance Score}\}$\\
Please wrap your response into a JSON object that contains keys `Clue of Step {}`\\
An example is given below. Please format your json output in a **LIST** form.
$\{\text{Example thought step}\}$
\end{tcolorbox}

\vspace{+5pt}

\subsection{Prompt for Code Generation}
\begin{tcolorbox}[colback=white!95!gray, colframe=black, width=1\textwidth, arc=4mm, boxrule=0.5mm]
You are a professional Python engineer. You will be given a problem desciption.\\
Please analyze the problem and generate executable python code to solve the problem.\\
The problem:\\
$\{\text{Problem Description}\}$\\
-----Clues-----\\
$\{\text{Current Thought Steps}\}$\\
-----Instruction-----\\
Now we have the clues above. Please refer to the clues and solve the problem. Please only output the code ONLY, no example usage or other explanation.
\end{tcolorbox}

\end{document}